\pdfoutput=1
\documentclass[article]{IEEEtran}
\usepackage{graphicx}
\usepackage[table,xcdraw]{xcolor}
\usepackage{adjustbox}
\usepackage{tabularx, booktabs}
\usepackage{tablefootnote}

\usepackage[space]{cite}

\usepackage{tikz}
\usepackage[cmex10]{amsmath}
\usepackage{amsfonts}
\usepackage{algorithm}
\usepackage[noend]{algpseudocode}
\usepackage{array}

\usepackage{mdwmath}

\usepackage{eqparbox}

\usepackage[]{caption}
\usepackage{hyperref}
\hypersetup{
    colorlinks=false,
    linkcolor=black,
    filecolor=magenta,      
    urlcolor=black,
    citecolor=blue,
}
\begin{document}
%
\title{LSTM Fully Convolutional Networks for Time Series Classification}
%
%
\makeatletter
\let\@fnsymbol\@arabic
\makeatother

\author{Fazle~Karim$^{1 *}$,
        Somshubra~Majumdar$^{2 *}$,
        Houshang~Darabi$^{1}$,~\IEEEmembership{Senior Member,~IEEE,
        and~Shun~Chen$^{1}$}
\thanks{$^{*}$Equal contribution.}
\thanks{$^{1}$Mechanical and Industrial Engineering, University of Illinois at Chicago, Chicago,IL}
\thanks{$^{2}$Computer Science, University of Illinois at Chicago, Chicago, IL}}
\maketitle

\begin{abstract}
Fully convolutional neural networks (FCN) have been shown to achieve state-of-the-art performance on the task of classifying time series sequences. We propose the augmentation of fully convolutional networks with long short term memory recurrent neural network (LSTM RNN) sub-modules for time series classification. Our proposed models significantly enhance the performance of fully convolutional networks with a nominal increase in model size and require minimal preprocessing of the dataset. The proposed Long Short Term Memory Fully Convolutional Network (LSTM-FCN) achieves state-of-the-art performance compared to others. We also explore the usage of attention mechanism to improve time series classification with the Attention Long Short Term Memory Fully Convolutional Network (ALSTM-FCN).  Utilization of the attention mechanism allows one to visualize the decision process of the LSTM cell. Furthermore, we propose fine-tuning as a method to enhance the performance of trained models. An  overall  analysis  of  the  performance  of  our  model  is  provided  and  compared  to  other techniques.
\end{abstract}

\begin{IEEEkeywords}
 Convolutional Neural Network, Long Short Term Memory Recurrent Neural Network, Time Series Classification
\end{IEEEkeywords}

%
\IEEEpeerreviewmaketitle

\section{Introduction}

\bstctlcite{IEEEexample:BSTcontrol}

Over the past decade, there has been an increased interest in time series classification. Time series data is ubiquitous, existing in weather readings, financial recordings, industrial observations, and psychological signals \cite{kadous2002temporal}. In this paper two deep learning models to classify time series datasets are proposed, both of which outperform existing state-of-the-art models.
\par A plethora of research have been done using feature-based approaches or methods to extract a set of features that represent time series patterns. Bag-of-Words (BoW) \cite{Lin_2007}, Bag-of-features (TSBF) \cite{Baydogan_2013}, Bag-of-SFA-Symbols (BOSS) \cite{Sch_fer_2014}, BOSSVS \cite{schafer2016scalable}, Word ExtrAction for time Series cLassification (WEASEL) \cite{Schafer_2017}, have obtained promising results in the field. Bag-of-words quantizes the extracted features and feeds the BoW into a classifier. TSBF extracts multiple subsequences of random local information, which a supervised learner condenses into a cookbook used to predict time series labels. BOSS introduces a combination of a distance based classifier and histograms. The histograms represent substructures of a time series that are created using a symbolic Fourier approximation. BOSSVS extends this method by proposing a vector space model to reduce time complexity while maintaining performance.  WEASEL converts time series into feature vectors using a sliding window. Machine learning algorithms utilize these feature vectors to detect and classify the time series.  All these classifiers require heavy feature extraction and feature engineering. \par Ensemble algorithms also yield state-of-the-art performance with time series classification problems. Three of the most successful ensemble algorithms that integrate various features of a time series are Elastic Ensemble (PROP) \cite{Lines_2014}, a model that integrates 11 time series classifiers using a weighted ensemble method, Shapelet ensemble (SE) \cite{bagnall2015time}, a model that applies a heterogeneous ensemble onto transformed “shapelets”, and a flat collective of transform based ensembles (COTE) \cite{bagnall2015time}, a model that fuses 35 various classifiers into a single classifier. 
\par Recently, deep neural networks have been employed for time series classification tasks. Multi-scale convolutional neural network (MCNN) \cite{cui2016multi}, fully convolutional network (FCN) \cite{wang2017time}, and residual network (ResNet) \cite{wang2017time} are deep learning approaches that take advantage of convolutional neural networks (CNN) for end-to-end classification of univariate time series. MCNN uses down-sampling, skip sampling and sliding window to preprocess the data. The performance of the MCNN classifier is highly dependent on the preprocessing applied to the dataset and the tuning of a large set of hyperparameters of that model. On the other hand, FCN and ResNet do not require any heavy preprocessing on the data or feature engineering. In this paper, we improve the performance of FCN by augmenting the FCN module with either a Long Short Term Recurrent Neural Network (LSTM RNN) sub-module , called LSTM-FCN, or a LSTM RNN  with attention, called ALSTM-FCN. Similar to FCN, both proposed models can be used to visualize the Class Activation Maps (CAM) of the convolutional layers to detect regions that contribute to the class label. In addition, the Attention LSTM can also be used detect regions of the input sequence that contribute to the class label through the context vector of the Attention LSTM cells.  A major advantages of the LSTM-FCN and ALSTM-FCN models is that it does not require heavy preprocessing or feature engineering. Results indicate the new proposed models, LSTM-FCN and ALSTM-FCN, dramatically improve performance on the University of California Riverside (UCR) Benchmark datasets \cite{UCRArchive}. LSTM-FCN and ALSTM-FCN produce better results than several state-of-the-art ensemble algorithms on a majority of the UCR Benchmark datasets. 
\par This paper proposes two deep learning models for end-to-end time series classification. The proposed models do not require heavy preprocessing on the data or feature engineering. Both the models are tested on all 85 UCR time series benchmarks and outperform most of the state-of-the-art models. The remainder of the paper is organized as follows. Section \ref{Background Works} reviews the background work. Section \ref{LSTMFCN} presents the architecture of the proposed models. Section \ref{Experiments} analyzes and discusses the experiments performed. Finally, conclusions are drawn in Section \ref{conclusion}.

\section{Background Works}
\label{Background Works}
\subsection{Temporal Convolutions}
The input to a Temporal Convolutional Network is generally a time series signal. As stated in \textit{Lea et al.}\cite{Lea_2016}, let $X_{t} \in \mathbb{R}^{F_0}$ be the input feature vector of length $F_0$ for time step $t$ for $0 < t \leq T$. Note that the time T may vary for each sequence, and we denote the number of time steps in each layer as $T_l$. The true action label for each frame is given by $y_t \in \{1, . . . , C\}$, where C is the number of classes.

Consider $L$ convolutional layers. We apply a set of 1D filters on each of these layers that capture how the input signals evolve over the course of an action. According to \textit{Lea et al.} \cite{Lea_2016}, the filters for each layer are parameterized by tensor $W^{(l)} \in \mathbb{R}^{F_l \times d \times F_{l-1}} $ and biases $b^{(l)} \in \mathbb{R}^{F_l}$, where $l \in \{1, . . . , L\}$ is the layer index and $d$ is the filter duration. For the $l$-th layer, the $i$-th component of the (unnormalized) activation ${\mathbf {\hat{E}}}^{(l)}_{t} \in \mathbb{R}^{F_{l}}$ is a function of the incoming (normalized) activation matrix $E^{(l-1)} \in \mathbb{R}^{F_{l-1} \times T_{l-1}}$ from the previous layer 
\begin{equation}
{\mathbf {\hat{E}}}_{i, t}^{(l)} = f\left(b_{i}^{(l)} + \sum_{t'=1}^{d} \left< W_{i, t', .}^{(l)}, E_{., t+d-t'}^{(l-1)} \right> \right)
\end{equation}
for each time $t$ where $f(\cdot)$ is a Rectified Linear Unit.

We use Temporal Convolutional Networks as a feature extraction module in a Fully Convolutional Network (FCN) branch. A basic convolution block consists of a convolution layer, followed by batch normalization \cite{ioffe2015batch}, followed by an activation function, which can be either a Rectified Linear Unit or a Parametric Rectified Linear Unit \cite{Trottier2016}.

\subsection{Recurrent Neural Networks}
\def\x{{\mathbf x}}
\def\L{{\cal L}}
Recurrent Neural Networks, often shortened to RNNs, are a class of neural networks which exhibit temporal behaviour due to directed connections between units of an individual layer. As reported by \textit{Pascanu et al.} \cite{pascanu2013construct}, recurrent neural networks maintain a hidden vector $\mathbf h$, which is updated at time step $t$ as follows:
 
\begin{equation}
	\mathbf h_t = \tanh(\mathbf W\mathbf h_{t-1} + \mathbf I\mathbf  \x_t),
\end{equation}
tanh is the hyperbolic tangent function, $\mathbf W$ is the recurrent weight matrix and $\mathbf I$ is a projection matrix. The hidden state $\mathbf h$ is  used to make a prediction

\begin{equation}
	\mathbf y_t = \text{softmax}(\mathbf W\mathbf h_{t-1}),
\end{equation}
softmax provides a normalized probability distribution over the possible classes, $\sigma$ is the logistic sigmoid function and $\mathbf W$ is a weight matrix. By using $\mathbf h$ as the input to another RNN, we can stack RNNs, creating deeper architectures 

\begin{equation}
	\mathbf h_t^{l} = \sigma(\mathbf W\mathbf h_{t-1}^{l} + \mathbf I\mathbf h_t^{l-1}).
\end{equation}

\subsection{Long Short-Term Memory RNNs}
\def\x{{\mathbf x}}

Long short-term memory recurrent neural networks are an improvement over the general recurrent neural networks, which possess a vanishing gradient problem. As stated in \textit{Hochreiter et al.}\cite{hochreiter1997long}, LSTM RNNs address the vanishing gradient problem commonly found in ordinary recurrent neural networks by incorporating gating functions into their state dynamics. At each time step, an LSTM maintains a hidden vector $\mathbf h$ and a memory vector $\mathbf m$  responsible for controlling state updates and outputs. More concretely, \textit{Graves et al.} \cite{graves2012supervised} define the computation at time step $t$ as follows :
\begin{equation}
	\begin{split}
		& \mathbf g^u = \sigma(\mathbf W^u\mathbf h_{t-1}  + \mathbf I^u\x_t ) \\
		& \mathbf g^f = \sigma(\mathbf W^f\mathbf h_{t-1} + \mathbf I^f\x_t) \\
		& \mathbf g^o = \sigma(\mathbf W^o\mathbf h_{t-1} + \mathbf I^o\x_t) \\
		& \mathbf g^c = \tanh(\mathbf W^c\mathbf h_{t-1} + \mathbf I^c\x_t) \\
		& \mathbf m_t = \mathbf g^f \odot \mathbf m_{t-1}\mathbf \ + \  \mathbf g^u \odot 
		\mathbf g^c \\
		& \mathbf h_t = \tanh(\mathbf g^o \odot \mathbf m_t) 
	\end{split}
\end{equation}
where $\sigma$ is the logistic sigmoid function, $\odot$ represents elementwise multiplication, $\mathbf W^u, \mathbf W^f, \mathbf W^o, \mathbf W^c$ are recurrent weight matrices and $\mathbf I^u, \mathbf I^f, \mathbf I^o, \mathbf I^c$ are projection matrices.

While LSTMs possess the ability to learn temporal dependencies in sequences, they have difficulty with long term dependencies in long sequences. The attention mechanism proposed by \textit{Bahdanau et al.} \cite{bahdanau2014neural} can help the LSTM RNN learn these dependencies. 

\subsection{Attention Mechanism}
The attention mechanism is a technique often used in neural translation of text, where a context vector $C$ is conditioned on the target sequence $y$. As discussed in \textit{Bahdanau et al.}\cite{bahdanau2014neural}, the context vector $c_i$ depends on a sequence of \textit{annotations} $(h_1, ..., h_{T_{x}})$ to which an encoder maps the input sequence. Each annotation $h_i$ contains information about the whole input sequence with a strong focus on the parts surrounding the $i$-th word of the input sequence.

The context vector $c_i$ is then computed as a weighted sum of these annotations $h_i$:
\begin{equation}
    c_i = \sum_{j=1}^{T_x} \alpha_{ij}h_j.
\end{equation}

The weight $\alpha_{ij}$ of each annotation $h_j$ is computed by : 
\begin{equation}
    \alpha_{ij} = \frac{exp(e_{ij})}{\sum_{k=1}^{T_x} exp(e_{ik})}
\end{equation}
where $e_{ij} = a(s_{i-1}, h_j)$ is an \textit{alignment model}, which scores how well the input around position $j$ and the output at position $i$ match. The score is based on the RNN hidden state $s_{i−1}$ and the $j$-th annotation $h_j$ of the input sentence. 

\textit{Bahdanau et al.}\cite{bahdanau2014neural} parametrize the alignment model $a$ as a feedforward neural network which is jointly trained with all the other components of the model. The alignment model directly computes a soft alignment, which allows the gradient of the cost function to be backpropagated. 

\begin{figure*}
\center
\fbox{
\includegraphics[width=0.85\linewidth]{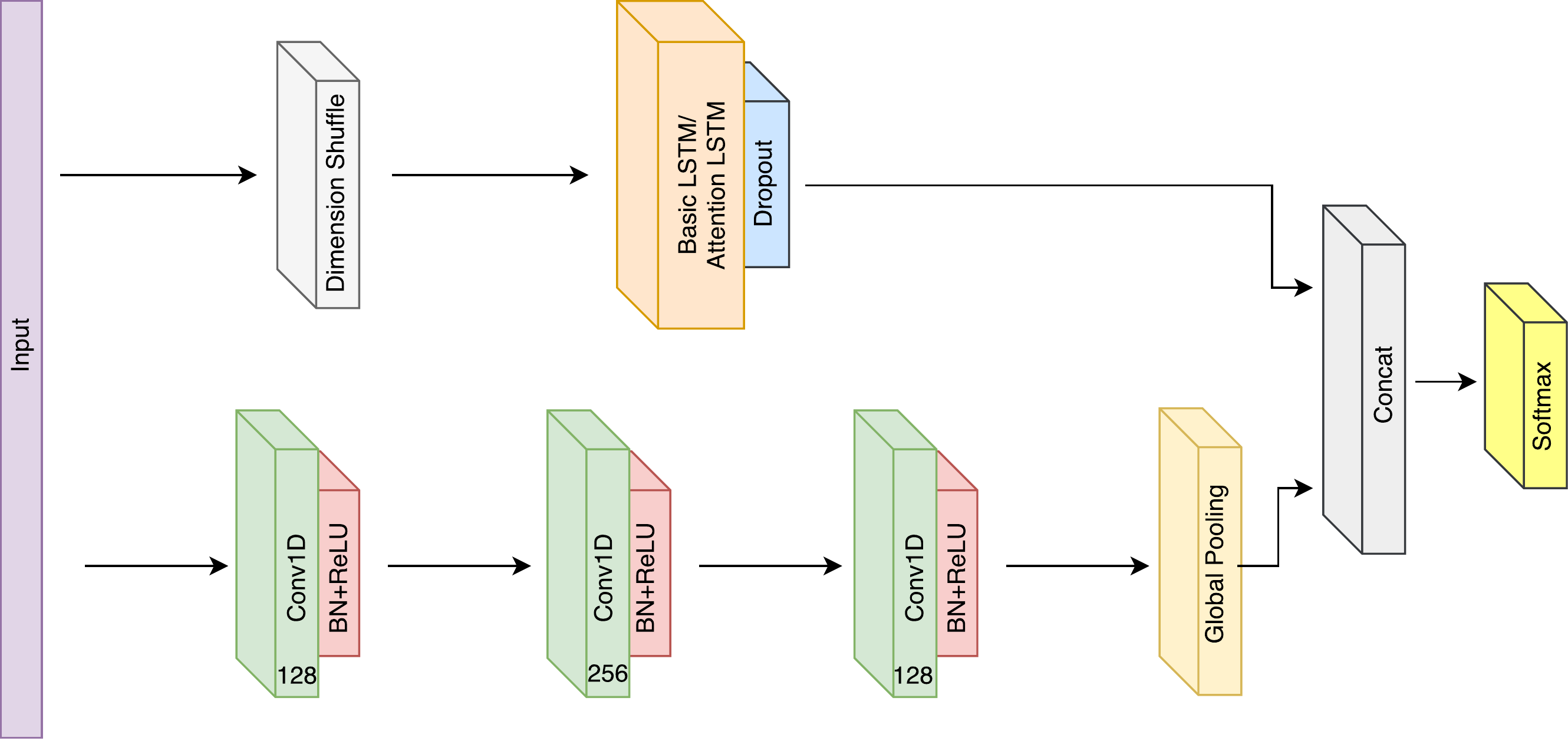}
}
\center
\caption{The LSTM-FCN architecture. LSTM cells can be replaced by Attention LSTM cells to construct the ALSTM-FCN architecture.}
\label{fig:arch}

\end{figure*}

\section{LSTM Fully Convolutional Network}
\label{LSTMFCN}
\subsection{Network Architecture}

Temporal convolutions have proven to be an effective learning model for time series classification problems \cite{wang2017time}. Fully Convolutional Networks comprised of temporal convolutions are typically used as feature extractors, and global average pooling \cite{lin2013network} is used to reduce the number of parameters in the model prior to classification. In the proposed models, the fully convolutional block is augmented by an LSTM block followed by dropout \cite{srivastava2014dropout}, as shown in Fig.\ref{fig:arch}. 

The fully convolutional block consists of three stacked temporal convolutional blocks with filter sizes of 128, 256, and 128 respectively. Each convolutional block is identical to the convolution block in the CNN architecture proposed by \textit{Wang et al.}\cite{wang2017time}. Each block consists of a temporal convolutional layer, which is accompanied by batch normalization \cite{ioffe2015batch} (momentum of $0.99$, epsilon of $0.001$) followed by a ReLU activation function. Finally, global average pooling is applied following the final convolution block. 

Simultaneously, the time series input is conveyed into a dimension shuffle layer (explained more in Section \ref{net_input}). The transformed time series from the dimension shuffle is then passed into the LSTM block. The LSTM block comprises of either a general LSTM layer or an Attention LSTM layer, followed by a dropout. The output of the global pooling layer and the LSTM block is concatenated and passed onto a softmax classification layer.

\subsection{Network Input}
\label{net_input}
The fully convolutional block and LSTM block perceive the same time series input in two different views. The fully convolutional block views the time series as a univariate time series with multiple time steps. If there is a time series of length $N$, the fully convolutional block will receive the data in $N$ time steps. 

Contrarily, the LSTM block in the proposed architecture receives the input time series as a multivariate time series with a single time step. This is accomplished by the dimension shuffle layer, which transposes the temporal dimension of the time series. A univariate time series of length $N$, after transformation, will be viewed as a multivariate time series (having $N$ variables) with a single time step. 

This approach is key to the enhanced performance of the proposed architecture. In contrast, when the LSTM block received the univariate time series with $N$ time steps, the performance was significantly reduced due to rapid overfitting on small short-sequence UCR datasets and a failure to learn long term dependencies in the larger long-sequence UCR datasets. 

\begin{figure*}
\center
\includegraphics[width=0.7\linewidth]{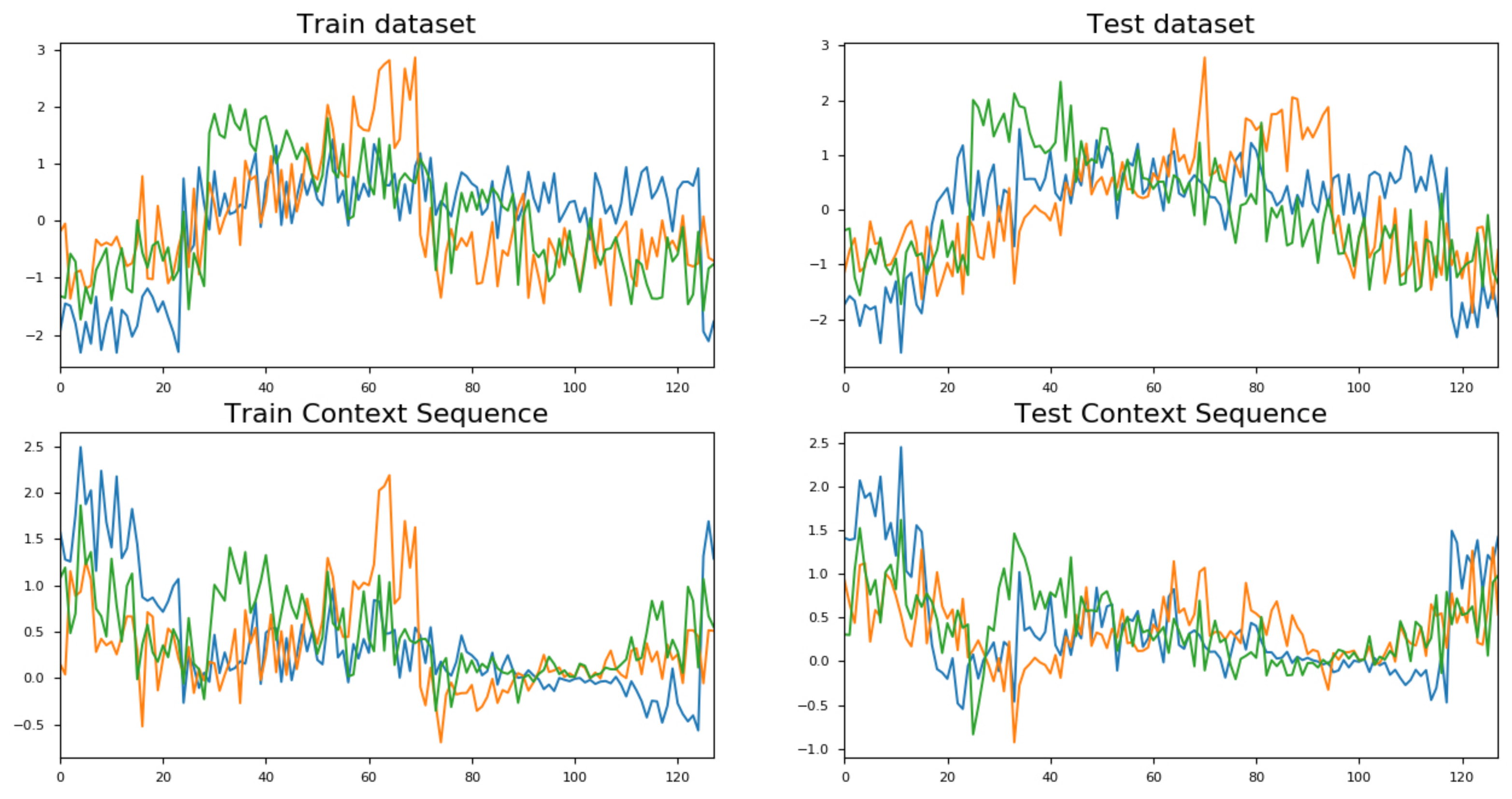}
\caption{Visualization of context vector on CBF dataset.}
\label{fig:attention_cbf}
\center
\end{figure*}

\subsection{Fine-Tuning of Models}
Transfer learning is a technique wherein the knowledge gained from training a model on a dataset can be reused when training the model on another dataset, such that the domain of the new dataset has some similarity with the prior domain \cite{yosinski2014transferable}. Similarly, \textit{fine-tuning} can be described as transfer learning on the same dataset.

The training procedure can thus be split into two distinct phases. In the \textit{initial phase}\label{phase1}, the optimal hyperparameters for the model are selected for a given dataset. The model is then trained on the given dataset with these hyperparameter settings. In the second step, we apply \textit{fine-tuning} to this initial model.

The procedure of transfer learning is iterated over in the \textit{fine-tuning phase}\label{phase2}, using the original dataset. Each repetition is initialized using the model weight of the previous iteration. At each iteration the learning rate is halved. Furthermore, the batch size is halved once every alternate iteration. This is done until the initial learning rate is $1e{-4}$ and batch size is 32. The procedure is repeated $K$ times, where K is an arbitrary constant, generally set as 5.

\begin{algorithm}
\begin{algorithmic}[1]
\caption{Fine-tuning}\label{finetuning}
\For {$i < K$}
\State $\textit{model}_{weights} \gets \textit{\textit{initial\_model}}_{weights}$
\State $\textit{Train(model, initial\_lr, batchsize)}$
\State $\textit{initial\_model}_{weights} \gets \textit{model}_{weights}$
\State $\textit{i} \gets \textit{i + 1}$
\State $\textit{initial\_lr} \gets updateLearningRate(\textit{initial\_lr}, \textit{i})$
\State $\textit{batchsize} \gets updateBatchsize(\textit{batchsize}, \textit{i})$
\EndFor
\end{algorithmic}
\end{algorithm}

\section{Experiments}
\label{Experiments}
The proposed models have been tested on all 85 UCR time series datasets \cite{UCRArchive}. The FCN block was kept constant throughout all experiments. The optimal number of LSTM cells was found by hyperparameter search over a range of 8 cells to 128 cells. The number of training epochs was generally kept constant at 2000 epochs, but was increased for datasets where the algorithm required a longer time to converge. Initial batch size of 128 was used, and halved for each successive iteration of the \textit{fine-tuning} algorithm. A high dropout rate of 80\% was used after the LSTM or Attention LSTM layer to combat overfitting. Class imbalance was handled via a class weighing scheme inspired by \textit{King et al.}\cite{king2001logistic}.

All models were trained via the Adam optimizer \cite{kingma2014adam}, with an initial learning rate of $1e{-3}$ and a final learning rate of $1e{-4}$. All convolution kernels were initialized with the initialization proposed by \textit{He et al.}\cite{he2015delving}. The learning rate was reduced by a factor of $1/{\sqrt[3]{2}}$ every 100 epochs of no improvement in the validation score, until the final learning rate was reached. No additional preprocessing was done on the UCR datasets as they have close to zero mean and unit variance. All models were \textit{fine-tuned}, and scores stated in Table \ref{tab:perf_tab} refer to the scores obtained by models prior to and after \textit{fine-tuning}. \footnote{The codes and weights of each models are available at \href{https://github.com/houshd/LSTM-FCN}{https://github.com/houshd/LSTM-FCN}}

\subsection{Evaluation Metrics}
In this paper, the proposed model was evaluated using accuracy, rank based statistics, and the mean per class error as stated by \textit{Wang et al.}\cite{wang2017time}.

The rank-based evaluations used are the arithmetic rank, geometric rank, and the Wilcoxon signed rank test. The arithmetic rank is the arithmetic mean of the rank of dataset. The geometric rank is the geometric mean of the rank of each dataset. The Wilcoxson signed rank test is used to compare the median rank of the proposed model and the existing state-of-the-art models. The null hypothesis and alternative hypothesis are as follows:
\begin{equation*}
H_o:Median_{\textit{proposed model}} = Median_{\textit{state-of-the-art model}}
\end{equation*}
\begin{equation*}
H_a:Median_{\textit{proposed model}} \neq Median_{\textit{state-of-the-art model}}
\end{equation*}
Mean Per Class Error (MPCE) is defined as the arithmetic mean of the per class error (PCE),
\begin{align*}
PCE_k &=\frac{1-accuracy}{\textit{number of unique classes}} \\
MPCE &=\frac{1}{\textit{K}} \sum{PCE_K}.
\end{align*}


\subsection{Results}

Fig. \ref{fig:attention_cbf} is an example of the visual representation of the Attention LSTM cell on the "CBF" dataset.  The points in the figure where the sequences are "squeezed" together are points at which all the classes have the same weight. These are the points in the time series at which the Attention LSTM can correctly identify the class. This is further supported by visual inspection of the actual time series. The squeeze points are points where each of the classes can be distinguished from each other, as shown in Fig. \ref{fig:attention_cbf}.

 \newcolumntype{C}{>{\centering\arraybackslash}X}
 \newcommand\mcxl[1]{\multicolumn{1}{|C|}{\bfseries #1}}
 \newcommand\mcx[1]{\multicolumn{1}{C }{\bfseries #1}}
 


 \begin{table}[]
 \centering
 \caption{Performance comparison of proposed models with the rest. }
\label{tab:perf_tab}
\begin{adjustbox}{width=1 \linewidth}

 \begin{tabularx}{0.62 \textwidth}{|C|C|C|C|C|C|}

    \hline
    Dataset & Existing SOTA \cite{Schafer_2017, wang2017time} & LSTM-FCN  & F-t LSTM-FCN & ALSTM-FCN & F-t ALSTM-FCN \\
    \hline
    Adiac & 0.8570 & \cellcolor[rgb]{ .776,  .878,  .706}0.8593 & \cellcolor[rgb]{ .776,  .878,  .706}0.8849 & \cellcolor[rgb]{ .776,  .878,  .706}0.8670 & \cellcolor[rgb]{ .776,  .878,  .706}0.8900*\\
    \hline
    ArrowHead & 0.8800 & \cellcolor[rgb]{ .776,  .878,  .706}0.9086 & \cellcolor[rgb]{ .776,  .878,  .706}0.9029 & \cellcolor[rgb]{ .776,  .878,  .706}0.9257* & \cellcolor[rgb]{ .776,  .878,  .706}0.9200 \\
    \hline
    Beef  & 0.9000 & \cellcolor[rgb]{ .776,  .878,  .706}0.9000 & \cellcolor[rgb]{ .776,  .878,  .706}0.9330 & \cellcolor[rgb]{ .776,  .878,  .706}0.9333* & \cellcolor[rgb]{ .776,  .878,  .706}0.9333* \\
    \hline
    BeetleFly & 0.9500 & 0.9500 & \cellcolor[rgb]{ .776,  .878,  .706}1.0000* & \cellcolor[rgb]{ .776,  .878,  .706}1.0000* & \cellcolor[rgb]{ .776,  .878,  .706}1.0000* \\
    \hline
    BirdChicken & 0.9500 & \cellcolor[rgb]{ .776,  .878,  .706}1.0000* & \cellcolor[rgb]{ .776,  .878,  .706}1.0000* & \cellcolor[rgb]{ .776,  .878,  .706}1.0000* & \cellcolor[rgb]{ .776,  .878,  .706}1.0000* \\
    \hline
    Car   & 0.9330 & \cellcolor[rgb]{ .776,  .878,  .706}0.9500 & \cellcolor[rgb]{ .776,  .878,  .706}0.9670 & \cellcolor[rgb]{ .776,  .878,  .706}0.9667 & \cellcolor[rgb]{ .776,  .878,  .706}0.9833* \\
    \hline
    CBF   & 1.0000 & 0.9978 & \cellcolor[rgb]{ .776,  .878,  .706}1.0000* & 0.9967 & 0.9967 \\
    \hline
    ChloConc & 0.8720 & 0.8099 & \cellcolor[rgb]{ .776,  .878,  .706}1.0000* & 0.8070 & 0.8070 \\
    \hline
    CinC\_ECG & 0.9949 & 0.8862 & 0.9094 & 0.9058 & 0.9058 \\
    \hline
    Coffee & 1.0000 & \cellcolor[rgb]{ .776,  .878,  .706}1.0000* & \cellcolor[rgb]{ .776,  .878,  .706}1.0000* & \cellcolor[rgb]{ .776,  .878,  .706}1.0000* & \cellcolor[rgb]{ .776,  .878,  .706}1.0000* \\
    \hline
    Computers & 0.8480 & \cellcolor[rgb]{ .776,  .878,  .706}0.8600 & \cellcolor[rgb]{ .776,  .878,  .706}0.8600 & \cellcolor[rgb]{ .776,  .878,  .706}0.8640* & \cellcolor[rgb]{ .776,  .878,  .706}0.8640* \\
    \hline
    Cricket\_X & 0.8210 & 0.8077 & \cellcolor[rgb]{ .776,  .878,  .706}0.8256* & 0.8051 & 0.8051 \\
    \hline
    Cricket\_Y & 0.8256 & 0.8179 & \cellcolor[rgb]{ .776,  .878,  .706}0.8256* & 0.8205 & 0.8205 \\
    \hline
    Cricket\_Z & 0.8154 & 0.8103 & \cellcolor[rgb]{ .776,  .878,  .706}0.8257 & \cellcolor[rgb]{ .776,  .878,  .706}0.8308 & \cellcolor[rgb]{ .776,  .878,  .706}0.8333* \\
    \hline
    DiaSizeRed & 0.9670 & \cellcolor[rgb]{ .776,  .878,  .706}0.9673 & \cellcolor[rgb]{ .776,  .878,  .706}0.9771* & \cellcolor[rgb]{ .776,  .878,  .706}0.9739 & \cellcolor[rgb]{ .776,  .878,  .706}0.9739 \\
    \hline
    DistPhxAgeGp & 0.8350 & \cellcolor[rgb]{ .776,  .878,  .706}0.8600 & \cellcolor[rgb]{ .776,  .878,  .706}0.8600 & \cellcolor[rgb]{ .776,  .878,  .706}0.8625* & \cellcolor[rgb]{ .776,  .878,  .706}0.8600 \\
    \hline
    DistPhxCorr & 0.8200 & \cellcolor[rgb]{ .776,  .878,  .706}0.8250 & \cellcolor[rgb]{ .776,  .878,  .706}0.8217 & \cellcolor[rgb]{ .776,  .878,  .706}0.8417* & \cellcolor[rgb]{ .776,  .878,  .706}0.8383 \\
    \hline
    DistPhxTW & 0.7900 & \cellcolor[rgb]{ .776,  .878,  .706}0.8175 & \cellcolor[rgb]{ .776,  .878,  .706}0.8100 & \cellcolor[rgb]{ .776,  .878,  .706}0.8175 & \cellcolor[rgb]{ .776,  .878,  .706}0.8200* \\
    \hline
    Earthquakes & 0.8010 & \cellcolor[rgb]{ .776,  .878,  .706}0.8354* & \cellcolor[rgb]{ .776,  .878,  .706}0.8261 & \cellcolor[rgb]{ .776,  .878,  .706}0.8292 & \cellcolor[rgb]{ .776,  .878,  .706}0.8292 \\
    \hline
    ECG200 & 0.9200 & 0.9000 & \cellcolor[rgb]{ .776,  .878,  .706}0.9200* & 0.9100 & \cellcolor[rgb]{ .776,  .878,  .706}0.9200 \\
    \hline
    ECG5000 & 0.9482 & 0.9473 & 0.9478 & \cellcolor[rgb]{ .776,  .878,  .706}0.9484 & \cellcolor[rgb]{ .776,  .878,  .706}0.9496* \\
    \hline
    ECGFiveDays & 1.0000 & 0.9919 & 0.9942 & 0.9954 & 0.9954 \\
    \hline
    ElectricDevices & 0.7993 & 0.7681 & 0.7633 & 0.7672 & 0.7672 \\
    \hline
    FaceAll & 0.9290 & \cellcolor[rgb]{ .776,  .878,  .706}0.9402 & \cellcolor[rgb]{ .776,  .878,  .706}0.9680 & \cellcolor[rgb]{ .776,  .878,  .706}0.9657 & \cellcolor[rgb]{ .776,  .878,  .706}0.9728* \\
    \hline
    FaceFour & 1.0000 & 0.9432 & 0.9772 & 0.9432 & 0.9432 \\
    \hline
    FacesUCR & 0.9580 & 0.9293 & \cellcolor[rgb]{ .776,  .878,  .706}0.9898* & 0.9434 & 0.9434 \\
    \hline
    FiftyWords & 0.8198 & 0.8044 & 0.8066 & \cellcolor[rgb]{ .776,  .878,  .706}0.8242 & \cellcolor[rgb]{ .776,  .878,  .706}0.8286* \\
    \hline
    Fish  & 0.9890 & 0.9829 & 0.9886 & 0.9771 & 0.9771 \\
    \hline
    FordA & 0.9727 & 0.9272 & \cellcolor[rgb]{ .776,  .878,  .706}0.9733* & 0.9267 & 0.9267 \\
    \hline
    FordB & 0.9173 & \cellcolor[rgb]{ .776,  .878,  .706}0.9180 & \cellcolor[rgb]{ .776,  .878,  .706}0.9186* & 0.9158 & 0.9158 \\
    \hline
    Gun\_Point & 1.0000 & \cellcolor[rgb]{ .776,  .878,  .706}1.0000* & \cellcolor[rgb]{ .776,  .878,  .706}1.0000* & \cellcolor[rgb]{ .776,  .878,  .706}1.0000* & \cellcolor[rgb]{ .776,  .878,  .706}1.0000* \\
    \hline
    Ham   & 0.7810 & 0.7714 & \cellcolor[rgb]{ .776,  .878,  .706}0.8000 & \cellcolor[rgb]{ .776,  .878,  .706}0.8381* & \cellcolor[rgb]{ .776,  .878,  .706}0.8000 \\
    \hline
    HandOutlines & 0.9487 & 0.8930 & 0.8870 & 0.9030 & 0.9030 \\
    \hline
    Haptics & 0.5510 & \cellcolor[rgb]{ .776,  .878,  .706}0.5747* & \cellcolor[rgb]{ .776,  .878,  .706}0.5584 & \cellcolor[rgb]{ .776,  .878,  .706}0.5649 & \cellcolor[rgb]{ .776,  .878,  .706}0.5584 \\
    \hline
    Herring & 0.7030 & \cellcolor[rgb]{ .776,  .878,  .706}0.7656* & \cellcolor[rgb]{ .776,  .878,  .706}0.7188 & \cellcolor[rgb]{ .776,  .878,  .706}0.7500 & \cellcolor[rgb]{ .776,  .878,  .706}0.7656* \\
    \hline
    InlineSkate & 0.6127 & 0.4655 & 0.5000 & 0.4927 & 0.4927 \\
    \hline
    InsWngSnd & 0.6525 & \cellcolor[rgb]{ .776,  .878,  .706}0.6616 & \cellcolor[rgb]{ .776,  .878,  .706}0.6696 & \cellcolor[rgb]{ .776,  .878,  .706}0.6823* & \cellcolor[rgb]{ .776,  .878,  .706}0.6818 \\
    \hline
    ItPwDmd & 0.9700 & 0.9631 & 0.9699 & 0.9602 & \cellcolor[rgb]{ .776,  .878,  .706}0.9708* \\
    \hline
    LrgKitApp & 0.8960 & \cellcolor[rgb]{ .776,  .878,  .706}0.9200* & \cellcolor[rgb]{ .776,  .878,  .706}0.9200* & \cellcolor[rgb]{ .776,  .878,  .706}0.9067 & \cellcolor[rgb]{ .776,  .878,  .706}0.9120 \\
    \hline
    Lighting2 & 0.8853 & 0.8033 & 0.8197 & 0.7869 & 0.7869 \\
    \hline
    Lighting7 & 0.8630 & 0.8356 & \cellcolor[rgb]{ .776,  .878,  .706}0.9178* & 0.8219 & \cellcolor[rgb]{ .776,  .878,  .706}0.9178* \\
    \hline
    Mallat & 0.9800 & \cellcolor[rgb]{ .776,  .878,  .706}0.9808 & \cellcolor[rgb]{ .776,  .878,  .706}0.9834 & \cellcolor[rgb]{ .776,  .878,  .706}0.9838 & \cellcolor[rgb]{ .776,  .878,  .706}0.9842* \\
    \hline
    Meat  & 1.0000 & 0.9167 & \cellcolor[rgb]{ .776,  .878,  .706}1.0000* & 0.9833 & \cellcolor[rgb]{ .776,  .878,  .706}1.0000* \\
    \hline
    MedicalImages & 0.7920 & \cellcolor[rgb]{ .776,  .878,  .706}0.8013 & \cellcolor[rgb]{ .776,  .878,  .706}0.8066* & \cellcolor[rgb]{ .776,  .878,  .706}0.7961 & \cellcolor[rgb]{ .776,  .878,  .706}0.7961 \\
    \hline
    MidPhxAgeGp & 0.8144 & 0.8125 & \cellcolor[rgb]{ .776,  .878,  .706}0.8150 & \cellcolor[rgb]{ .776,  .878,  .706}0.8175* & 0.8075 \\
    \hline
    MidPhxCorr & 0.8076 & \cellcolor[rgb]{ .776,  .878,  .706}0.8217 & \cellcolor[rgb]{ .776,  .878,  .706}0.8333 & \cellcolor[rgb]{ .776,  .878,  .706}0.8400 & \cellcolor[rgb]{ .776,  .878,  .706}0.8433* \\
    \hline
    MidPhxTW & 0.6120 & \cellcolor[rgb]{ .776,  .878,  .706}0.6165 & \cellcolor[rgb]{ .776,  .878,  .706}0.6466 & \cellcolor[rgb]{ .776,  .878,  .706}0.6466* & \cellcolor[rgb]{ .776,  .878,  .706}0.6316 \\
    \hline
    MoteStrain & 0.9500 & 0.9393 & \cellcolor[rgb]{ .776,  .878,  .706}0.9569* & 0.9361 & 0.9361 \\
    \hline
    NonInv\_Thor1 & 0.9610 & \cellcolor[rgb]{ .776,  .878,  .706}0.9654 & \cellcolor[rgb]{ .776,  .878,  .706}0.9657 & \cellcolor[rgb]{ .776,  .878,  .706}0.9751 & \cellcolor[rgb]{ .776,  .878,  .706}0.9756* \\
    \hline
    NonInv\_Thor2 & 0.9550 & \cellcolor[rgb]{ .776,  .878,  .706}0.9623 & \cellcolor[rgb]{ .776,  .878,  .706}0.9613 & \cellcolor[rgb]{ .776,  .878,  .706}0.9664 & \cellcolor[rgb]{ .776,  .878,  .706}0.9674* \\
    \hline
    OliveOil & 0.9333 & 0.8667 & \cellcolor[rgb]{ .776,  .878,  .706}0.9333 & \cellcolor[rgb]{ .776,  .878,  .706}0.9333 & \cellcolor[rgb]{ .776,  .878,  .706}0.9667* \\
    \hline
    OSULeaf & 0.9880 & \cellcolor[rgb]{ .776,  .878,  .706}0.9959* & \cellcolor[rgb]{ .776,  .878,  .706}0.9959* & \cellcolor[rgb]{ .776,  .878,  .706}0.9959* & \cellcolor[rgb]{ .776,  .878,  .706}0.9917 \\
    \hline
    PhalCorr & 0.8300 & \cellcolor[rgb]{ .776,  .878,  .706}0.8368 & \cellcolor[rgb]{ .776,  .878,  .706}0.8392* & \cellcolor[rgb]{ .776,  .878,  .706}0.8380 & \cellcolor[rgb]{ .776,  .878,  .706}0.8357 \\
    \hline
    Phoneme & 0.3492 & \cellcolor[rgb]{ .776,  .878,  .706}0.3776* & \cellcolor[rgb]{ .776,  .878,  .706}0.3602 & \cellcolor[rgb]{ .776,  .878,  .706}0.3671 & \cellcolor[rgb]{ .776,  .878,  .706}0.3623 \\
    \hline
    Plane & 1.0000 & \cellcolor[rgb]{ .776,  .878,  .706}1.0000* & \cellcolor[rgb]{ .776,  .878,  .706}1.0000* & \cellcolor[rgb]{ .776,  .878,  .706}1.0000* & \cellcolor[rgb]{ .776,  .878,  .706}1.0000* \\
    \hline
    ProxPhxAgeGp & 0.8832 & \cellcolor[rgb]{ .776,  .878,  .706}0.8927* & \cellcolor[rgb]{ .776,  .878,  .706}0.8878 & \cellcolor[rgb]{ .776,  .878,  .706}0.8878 & \cellcolor[rgb]{ .776,  .878,  .706}0.8927* \\
    \hline
    ProxPhxCorr & 0.9180 & \cellcolor[rgb]{ .776,  .878,  .706}0.9450* & \cellcolor[rgb]{ .776,  .878,  .706}0.9313 & \cellcolor[rgb]{ .776,  .878,  .706}0.9313 & \cellcolor[rgb]{ .776,  .878,  .706}0.9381 \\
    \hline
    ProxPhxTW & 0.8150 & \cellcolor[rgb]{ .776,  .878,  .706}0.8350 & \cellcolor[rgb]{ .776,  .878,  .706}0.8275 & \cellcolor[rgb]{ .776,  .878,  .706}0.8375* & \cellcolor[rgb]{ .776,  .878,  .706}0.8375* \\
    \hline
    RefDev & 0.5813 & \cellcolor[rgb]{ .776,  .878,  .706}0.5813 & \cellcolor[rgb]{ .776,  .878,  .706}0.5947* & \cellcolor[rgb]{ .776,  .878,  .706}0.5840 & \cellcolor[rgb]{ .776,  .878,  .706}0.5840 \\
    \hline
    ScreenType & 0.7070 & 0.6693 & \cellcolor[rgb]{ .776,  .878,  .706}0.7073 & 0.6907 & 0.6907 \\
    \hline
    ShapeletSim & 1.0000 & 0.9722 & \cellcolor[rgb]{ .776,  .878,  .706}1.0000* & 0.9833 & 0.9833 \\
    \hline
    ShapesAll & 0.9183 & 0.9017 & 0.9150 & \cellcolor[rgb]{ .776,  .878,  .706}0.9183 & \cellcolor[rgb]{ .776,  .878,  .706}0.9217* \\
    \hline
    SmlKitApp & 0.8030 & \cellcolor[rgb]{ .776,  .878,  .706}0.8080 & \cellcolor[rgb]{ .776,  .878,  .706}0.8133* & 0.7947 & \cellcolor[rgb]{ .776,  .878,  .706}0.8133* \\
    \hline
    SonyAIBOI & 0.9850 & 0.9817 & \cellcolor[rgb]{ .776,  .878,  .706}0.9967 & 0.9700 & \cellcolor[rgb]{ .776,  .878,  .706}0.9983* \\
    \hline
    SonyAIBOII & 0.9620 & \cellcolor[rgb]{ .776,  .878,  .706}0.9780 & \cellcolor[rgb]{ .776,  .878,  .706}0.9822* & \cellcolor[rgb]{ .776,  .878,  .706}0.9748 & \cellcolor[rgb]{ .776,  .878,  .706}0.9790 \\
    \hline
    StarlightCurves & 0.9796 & 0.9756 & 0.9763 & 0.9767 & 0.9767 \\
    \hline
    Strawberry & 0.9760 & \cellcolor[rgb]{ .776,  .878,  .706}0.9838 & \cellcolor[rgb]{ .776,  .878,  .706}0.9864 & \cellcolor[rgb]{ .776,  .878,  .706}0.9838 & \cellcolor[rgb]{ .776,  .878,  .706}0.9865* \\
    \hline
    SwedishLeaf & 0.9664 & \cellcolor[rgb]{ .776,  .878,  .706}0.9792 & \cellcolor[rgb]{ .776,  .878,  .706}0.9840 & \cellcolor[rgb]{ .776,  .878,  .706}0.9856* & \cellcolor[rgb]{ .776,  .878,  .706}0.9856* \\
    \hline
    Symbols & 0.9668 & \cellcolor[rgb]{ .776,  .878,  .706}0.9839 & \cellcolor[rgb]{ .776,  .878,  .706}0.9849 & \cellcolor[rgb]{ .776,  .878,  .706}0.9869 & \cellcolor[rgb]{ .776,  .878,  .706}0.9889* \\
    \hline
    Synth\_Cntr & 1.0000 & 0.9933 & \cellcolor[rgb]{ .776,  .878,  .706}1.0000* & 0.9900 & 0.9900 \\
    \hline
    ToeSeg1 & 0.9737 & \cellcolor[rgb]{ .776,  .878,  .706}0.9825 & \cellcolor[rgb]{ .776,  .878,  .706}0.9912* & \cellcolor[rgb]{ .776,  .878,  .706}0.9868 & \cellcolor[rgb]{ .776,  .878,  .706}0.9868 \\
    \hline
    ToeSeg2 & 0.9615 & 0.9308 & 0.9462 & 0.9308 & 0.9308 \\
    \hline
    Trace & 1.0000 & \cellcolor[rgb]{ .776,  .878,  .706}1.0000* & \cellcolor[rgb]{ .776,  .878,  .706}1.0000* & \cellcolor[rgb]{ .776,  .878,  .706}1.0000* & \cellcolor[rgb]{ .776,  .878,  .706}1.0000* \\
    \hline
    Two\_Patterns & 1.0000 & 0.9968 & 0.9973 & 0.9968 & 0.9968 \\
    \hline
    TwoLeadECG & 1.0000 & 0.9991 & \cellcolor[rgb]{ .776,  .878,  .706}1.0000* & 0.9991 & \cellcolor[rgb]{ .776,  .878,  .706}1.0000* \\
    \hline
    uWavGest\_X & 0.8308 & \cellcolor[rgb]{ .776,  .878,  .706}0.8490 & \cellcolor[rgb]{ .776,  .878,  .706}0.8498 & \cellcolor[rgb]{ .776,  .878,  .706}0.8481 & \cellcolor[rgb]{ .776,  .878,  .706}0.8504* \\
    \hline
    uWavGest\_Y & 0.7585 & \cellcolor[rgb]{ .776,  .878,  .706}0.7672* & \cellcolor[rgb]{ .776,  .878,  .706}0.7661 & \cellcolor[rgb]{ .776,  .878,  .706}0.7658 & \cellcolor[rgb]{ .776,  .878,  .706}0.7644 \\
    \hline
    uWavGest\_Z & 0.7725 & \cellcolor[rgb]{ .776,  .878,  .706}0.7973 & \cellcolor[rgb]{ .776,  .878,  .706}0.7993 & \cellcolor[rgb]{ .776,  .878,  .706}0.7982 & \cellcolor[rgb]{ .776,  .878,  .706}0.8007* \\
    \hline
    uWavGestAll & 0.9685 & 0.9618 & 0.9609 & 0.9626 & 0.9626 \\
    \hline
    Wafer & 1.0000 & 0.9992 & \cellcolor[rgb]{ .776,  .878,  .706}1.0000* & 0.9981 & 0.9981 \\
    \hline
    Wine  & 0.8890 & 0.8704 & \cellcolor[rgb]{ .776,  .878,  .706}0.8890 & \cellcolor[rgb]{ .776,  .878,  .706}0.9074* & \cellcolor[rgb]{ .776,  .878,  .706}0.9074* \\
    \hline
    WordsSynonyms & 0.7790 & 0.6708 & 0.6991 & 0.6677 & 0.6677 \\
    \hline
    Worms & 0.8052 & 0.6685 & 0.6851 & 0.6575 & 0.6575 \\
    \hline
    WormsTwoClass & 0.8312 & 0.7956 & 0.8066 & 0.8011 & 0.8011 \\
    \hline
    yoga  & 0.9183 & 0.9177 & 0.9163 & \cellcolor[rgb]{ .776,  .878,  .706}0.9190 & \cellcolor[rgb]{ .776,  .878,  .706}0.9237* \\
    \hline
    Count & -     & 43 & 65 & 51 & 57 \\
    \hline
    MPCE  & -     & 0.0318 & 0.0283 & 0.0301 & 0.0294 \\
    \hline
    Arith. Mean & -     & -     & 2.1529 & -     & 2.5647 \\
    \hline
    Geom. Mean & -     & -     & 1.8046 & -     & 1.8506 \\
    \hline

    \end{tabularx}%
    \end{adjustbox}
  \label{tab:addlabel}%

\tablefootnote{Green cells designate instances where our performance matches or exceeds state-of-the-art results. \textbf{*} denotes model with best performance.}
\end{table}%
 
\def\tabularxcolumn#1{m{#1}}

\newcolumntype{b}{>{\centering \arraybackslash\hsize=1.5\hsize}X}
\newcolumntype{s}{>{\centering \arraybackslash \hsize=.5\hsize}X}
\newcommand{\heading}[1]{\multicolumn{1}{c}{#1}}

\begin{table*}[]
\centering
\caption{Wilcoxon Signed Rank Test comparison of each Model}
\label{tab:pvaltab}
\begin{adjustbox}{width=1 \textwidth}

\begin{tabularx}{1.27 \textwidth}{|b|s|s|s|s|s|s|s|s|s|s|s|s|s|s|s|s|}

    \hline
          & WEASEL & 1-NN DTW CV & 1-NN DTW & BOSS  & Learning Shapelet & TSBF  & ST    & EE    & COTE  & MLP   & CNN   & ResNet & LSTM-FCN & F-t LSTM-FCN & ALSTM-FCN\\

    \hline
    WEASEL &       &       &       &       &       &       &       &       &       &       &       &       &       &       &  \\
    \hline
    1-NN DTW CV & 2.39E-10 &       &       &       &       &       &       &       &       &       &       &       &       &       &  \\
    \hline
    1-NN DTW & 2.53E-12 & 7.20E-04 &       &       &       &       &       &       &       &       &       &       &       &       &  \\
    \hline
    BOSS  & 4.27E-03 & 1.82E-07 & 5.31E-11 &       &       &       &       &       &       &       &       &       &       &       &  \\
    \hline
    Learning Shapelet & 2.00E-04 & 2.53E-02 & 2.33E-04 & 1.94E-02 &       &       &       &       &       &       &       &       &       &       &  \\
    \hline
    TSBF  & 2.18E-05 & \cellcolor[rgb]{ 1,  .78,  .808}\textcolor[rgb]{ .612,  0,  .024}{1.59E-01} & 2.49E-03 & 4.36E-03 & \cellcolor[rgb]{ 1,  .78,  .808}\textcolor[rgb]{ .612,  0,  .024}{4.73E-01} &       &       &       &       &       &       &       &       &       &  \\
    \hline
    ST    & \cellcolor[rgb]{ 1,  .78,  .808}\textcolor[rgb]{ .612,  0,  .024}{1.29E-01} & 1.05E-07 & 9.64E-11 & \cellcolor[rgb]{ 1,  .78,  .808}\textcolor[rgb]{ .612,  0,  .024}{2.39E-01} & 1.61E-03 & 3.60E-04 &       &       &       &       &       &       &       &       &  \\
    \hline
    EE    & 4.51E-05 & 3.45E-07 & 1.31E-10 & 1.37E-02 & \cellcolor[rgb]{ 1,  .78,  .808}\textcolor[rgb]{ .612,  0,  .024}{6.13E-01} & \cellcolor[rgb]{ 1,  .78,  .808}\textcolor[rgb]{ .612,  0,  .024}{2.02E-01} & 1.39E-03 &       &       &       &       &       &       &       &  \\
    \hline
    COTE  & \cellcolor[rgb]{ 1,  .78,  .808}\textcolor[rgb]{ .612,  0,  .024}{5.44E-01} & 3.05E-14 & 3.03E-16 & 6.21E-04 & 4.76E-07 & 1.13E-06 & 4.24E-03 & 3.54E-11 &       &       &       &       &       &       &  \\
    \hline
    MLP   & 2.56E-07 & \cellcolor[rgb]{ 1,  .78,  .808}\textcolor[rgb]{ .612,  0,  .024}{5.21E-01} & \cellcolor[rgb]{ 1,  .78,  .808}\textcolor[rgb]{ .612,  0,  .024}{3.41E-01} & 6.89E-05 & 1.44E-02 & \cellcolor[rgb]{ 1,  .78,  .808}\textcolor[rgb]{ .612,  0,  .024}{8.37E-02} & 6.76E-06 & 4.88E-03 & 2.84E-08 &       &       &       &       &       &  \\
    \hline
    FCN   & \cellcolor[rgb]{ 1,  .78,  .808}\textcolor[rgb]{ .612,  0,  .024}{2.77E-01} & 1.84E-10 & 2.14E-15 & 1.03E-03 & 3.65E-06 & 1.54E-06 & 8.85E-03 & 6.07E-06 & \cellcolor[rgb]{ 1,  .78,  .808}\textcolor[rgb]{ .612,  0,  .024}{4.82E-01} & 2.79E-09 &       &       &       &       &  \\
    \hline
    ResNet & \cellcolor[rgb]{ 1,  .78,  .808}\textcolor[rgb]{ .612,  0,  .024}{5.67E-01} & 1.82E-10 & 5.95E-15 & 4.38E-03 & 1.32E-05 & 3.56E-06 & 2.47E-02 & 1.09E-05 & \cellcolor[rgb]{ 1,  .78,  .808}\textcolor[rgb]{ .612,  0,  .024}{9.61E-01} & 4.64E-08 & \cellcolor[rgb]{ 1,  .78,  .808}\textcolor[rgb]{ .612,  0,  .024}{2.52E-01} &       &       &       &  \\
    \hline
    LSTM-FCN & 4.92E-06 & 1.92E-17 & 8.59E-21 & 3.00E-11 & 2.65E-12 & 4.04E-12 & 9.93E-13 & 5.14E-13 & 1.60E-07 & 1.61E-14 & 1.05E-07 & 4.91E-10 &       &       &  \\
    \hline
    F-t LSTM-FCN & 1.23E-08 & 5.17E-19 & 5.77E-22 & 3.35E-13 & 2.20E-13 & 1.12E-13 & 3.44E-14 & 1.25E-14 & 2.81E-10 & 5.09E-16 & 3.35E-12 & 4.58E-15 & 7.53E-05 &       &  \\
    \hline
    ALSTM-FCN & 1.34E-07 & 2.74E-18 & 5.14E-21 & 1.34E-12 & 3.38E-12 & 7.48E-13 & 7.11E-14 & 1.26E-13 & 1.30E-08 & 1.70E-15 & 3.74E-09 & 1.33E-11 & 8.53E-04 & 3.06E-02 &  \\
    \hline
    F-t ALSTM-FCN & 4.58E-08 & 1.01E-18 & 1.18E-21 & 1.44E-12 & 2.41E-12 & 4.63E-13 & 3.96E-14 & 4.12E-14 & 2.56E-09 & 1.87E-15 & 2.60E-10 & 1.12E-12 & 5.96E-05 & \cellcolor[rgb]{ 1,  .78,  .808}\textcolor[rgb]{ .612,  0,  .024}{1.89E-01} & \cellcolor[rgb]{ 1,  .78,  .808}\textcolor[rgb]{ .612,  0,  .024}{5.40E-02} \\
    \hline

\end{tabularx}
\end{adjustbox}

\end{table*}

\begin{figure}
\fbox{
\includegraphics[width=0.95\linewidth]{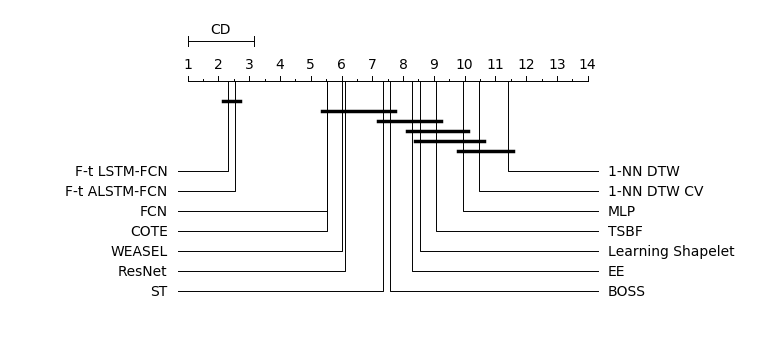}
}
\caption{Critical difference diagram of the arithmetic means of the ranks}
\label{fig:critical_diff}

\end{figure}

The performance of the proposed models on the UCR datasets are summarized in Table \ref{tab:perf_tab}. The colored cells are cells that outperform the state-of-the-art model for that dataset. Both proposed models, the ALSTM-FCN model and the LSTM-FCN model, with both phases, without \textit{fine-tuning} (Phase 1) and with \textit{fine-tuning} (Phase 2), outperforms the state-of-the-art models in at least 43 datasets. The average arithmetic rank in Fig. \ref{fig:critical_diff} indicates the superiority of our proposed models over the existing state-of-the-art models. This is further validated using the Wilcoxon signed rank test, where the p-value of each of the proposed models are less than 0.05 when compared to existing state-of-the-art models, Table \ref{tab:pvaltab}.

The Wilcoxon Signed Test also provides evidence that \textit{fine-tuning} maintains or improves the overall accuracy on each of the proposed models. The MPCE of the LSTM-FCN and ALSTM-FCN models was found to reduce by 0.0035 and 0.0007 respectively when \textit{fine-tuning} was applied. \textit{Fine-tuning} improves the accuracy of the LSTM-FCN models on a greater number of datasets as compared to the ALSTM-FCN models. We postulate that this discrepancy is due to the fact that the LSTM-FCN model contains fewer total parameters than the ALSTM-FCN model. This indicates a lower rate of overfitting on the UCR datasets. As a consequence, \textit{fine-tuning} is more effective on the LSTM-FCN models for the UCR datasets. 

A significant drawback of  \textit{fine-tuning} is that it requires more training time due to the added computational complexity of re-training the model using smaller batch sizes. The disadvantages of \textit{fine-tuning} are mitigated when using the ALSTM-FCN within Phase 1. At the end of Phase 1, the ALSTM-FCN model outperforms the Phase 1 LSTM-FCN model. One of the major advantage of using the Attention LSTM cell is it provides a visual representation of the attention vector. The Attention LSTM also benefits from \textit{fine-tuning}, but the effect is less significant as compared to the general LSTM model. A summary of the performance of each model type on certain characteristics is provided on Table \ref{tab:sum}.


\def\checkmark{\tikz\fill[scale=0.4](0,.35) -- (.25,0) -- (1,.7) -- (.25,.15) -- cycle;}

 \begin{table}[]
 \centering
 \caption{Summary of advantages of the proposed models}
\label{tab:sum}
\begin{adjustbox}{width=1 \linewidth}

 \begin{tabularx}{0.62 \textwidth}{|C|C|C|C|C|}
    \hline
    Advantage & LSTM-FCN & F-t LSTM-FCN & ALSTM-FCN & F-t ALSTM-FCN \\
    \hline
    Performance &       & \checkmark     &       & \checkmark \\
    \hline
    Visualization &       &       & \checkmark     & \checkmark \\
    \hline
    \end{tabularx}
  \end{adjustbox}
\end{table}

\section{Conclusion \& Future Work}
\label{conclusion}
With the proposed models, we achieve a potent improvement in the current state-of-the-art for time series classification using deep neural networks. Our baseline models, with and without fine-tuning, are trainable end-to-end with nominal preprocessing and are able to achieve significantly improved performance. LSTM-FCNs are able to augment FCN models, appreciably increasing their performance with a nominal increase in the number of parameters. ALSTM-FCNs provide one with the ability to visually inspect the decision process of the LSTM RNN and provide a strong baseline on their own. \textit{Fine-tuning} can be applied as a general procedure to a model to further elevate its performance. The strong increase in performance in comparison to the FCN models shows that LSTM RNNs can beneficially supplement the performance of FCN modules for time series classification. An overall analysis of the performance of our model is provided and compared to other techniques. 

There is further research to be done on understanding why the attention LSTM cell is unsuccessful in matching the performance of the general LSTM cell on some of the datasets. Furthermore, extension of the proposed models to multivariate time series is elementary, but has not been explored in this work.


%




\ifCLASSOPTIONcaptionsoff
  \newpage
\fi



\bibliographystyle{IEEEtran}

\bibliography{biblio}{}
%



%





\end{document}